\def\year{2022}\relax
\documentclass[letterpaper]{article} 
\usepackage{aaai22}  
\usepackage{times}  
\usepackage{helvet}  
\usepackage{courier}  
\usepackage[hyphens]{url}  
\usepackage{graphicx} 
\usepackage{amsmath}
\usepackage{booktabs}
\usepackage{multirow}
\usepackage{colortbl}
\usepackage{amssymb}
\usepackage{lipsum}
\usepackage{hyperref}
\usepackage[dvipsnames]{xcolor}
\definecolor{Highlight}{HTML}{39b54a}  
\definecolor{Bad}{HTML}{a8a8a8}  
\definecolor{Bg}{HTML}{e0f1ff}  

\usepackage{natbib}  
\usepackage{caption} 
\DeclareCaptionStyle{ruled}{labelfont=normalfont,labelsep=colon,strut=off} 
\usepackage{algorithm}
\usepackage{algorithmic}

\usepackage{newfloat}
\usepackage{listings}
\floatstyle{ruled}
\newfloat{listing}{tb}{lst}{}
\floatname{listing}{Listing}
\newcommand\blfootnote[1]{%
  \begingroup
  \renewcommand\thefootnote{}\footnote{#1}%
  \setcounter{footnote}{0}%
  \endgroup
}

\title{When Shift Operation Meets Vision Transformer: \\ An Extremely Simple Alternative to Attention Mechanism}
\author{Guangting Wang\textsuperscript{\rm 1}\equalcontrib, Yucheng Zhao\textsuperscript{\rm 1}\equalcontrib, Chuanxin Tang\textsuperscript{\rm 2}\equalcontrib, Chong Luo\textsuperscript{\rm 2} \blfootnote{This work was done during the internship of Guangting and Yucheng at MSRA}, Wenjun Zeng\textsuperscript{\rm 2}}
\affiliations{
    \textsuperscript{\rm 1} University of Science and Technology of China \\
    \textsuperscript{\rm 2} Microsoft Research Asia
}

\usepackage{bibentry}

\begin{document}
\maketitle

\begin{abstract}
   
   Attention mechanism has been widely believed as the key to success of vision transformers (ViTs), since it provides a flexible and powerful way to model spatial relationships. However, is the attention mechanism truly an indispensable part of ViT? Can it be replaced by some other alternatives? To demystify the role of attention mechanism, we simplify it into an extremely simple case: ZERO FLOP and ZERO parameter. Concretely, we revisit the shift operation. It does not contain any parameter or arithmetic calculation. The only operation is to exchange a small portion of the channels between neighboring features. Based on this simple operation, we construct a new backbone network, namely ShiftViT, where the attention layers in ViT are substituted by shift operations. Surprisingly, ShiftViT works quite well in several mainstream tasks, e.g., classification, detection, and segmentation. The performance is on par with or even better than the strong baseline Swin Transformer. These results suggest that the attention mechanism might not be the vital factor that makes ViT successful. It can be even replaced by a zero-parameter operation. We should pay more attentions to the remaining parts of ViT in the future work. Code is available at \href{github.com/microsoft/SPACH}{github.com/microsoft/SPACH}.

\end{abstract}

\section{Introduction}

Designing backbone networks plays a fundamental role in computer vision. Since the revolutionary progress of AlexNet \cite{AlexNet}, convolution neural networks (CNNs) have dominated this area for nearly 10 years. However, the recently developed Vision Transformers (ViTs) have shown potential to challenge this throne. The advantage of ViT was first demonstrated in image classification task \cite{ViT}, where the ViT backbone outperforms its CNN counterparts by a remarkable margin. Thanks to the promising results, the flourish of ViT variants rapidly broadcasts to many other computer vision tasks, such as object detection, semantic segmentation, and action recognition.

Despite the impressive performances of recent ViT variants, it is still not yet clear what makes ViT good for visual recognition tasks. Some conventional wisdom leans to credit the success to the attention mechanism, since it provides a flexible and powerful way to model spatial relationships. Concretely, the attention mechanism leverages a self-attention matrix to aggregate features from arbitrary locations. Compared with the convolution operation in CNN, it has two significant strengths. First, this mechanism opens a possibility to simultaneously capture both short- and long-ranged dependencies, and get rid of the local restriction of the convolution. Second, the interaction between two spatial locations dynamically depends on their own features, rather than a fixed convolutional kernel. Due to such good properties, some pieces of work believe it is the attention mechanism that facilitates the powerful expressive ability of ViTs.

\begin{figure}[t]
\centering
\includegraphics[width=\linewidth]{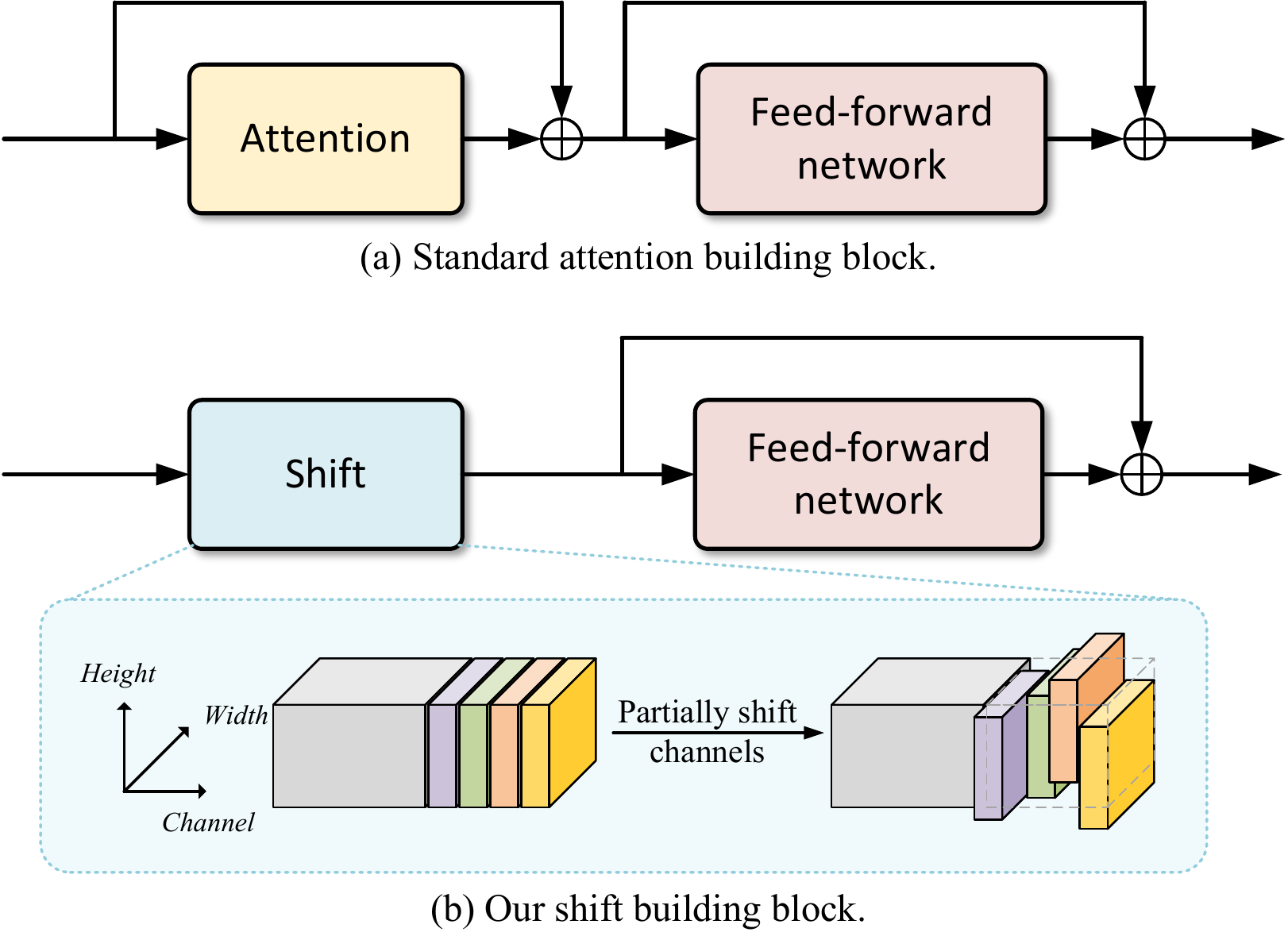}
\caption{An illustration of our shift building block. We propose to replace the attention layer with a simple shift operation in vision transformers. It spatially shifts a small portion of the channels along four directions, and the rest of the channels remain unchanged.}
\label{fig:architecture}
\end{figure}

However, are these two advantages truly the key to success? The answer is probably NOT. Some existing work proves that, even without these properties, the ViT variants can still work well. For the first one, the fully-global dependencies may not be inevitable. More and more ViTs introduce a local attention mechanism to restrict their attention scope within a small local region, e.g., Swin Transformer \cite{Swin} and Local ViT \cite{LocalGlobalTransformer}. The experiments show that the performance does not drop due to the local restriction. Besides, another line of research investigates the necessity of the dynamic aggregation. MLP-Mixer \cite{MLPMixer} proposes to substitute the attention layer with a linear projection layer, where the linear weights are not dynamically generated. In this case, it can still reach a leading performance on the ImageNet dataset.

Now that both global and dynamic properties might not be crucial for the ViT framework, what is the essential reason for the success of ViT? To figure it out, we further simplify the attention layer into an extremely simple case: \textit{NO global scope, NO dynamics, and even NO parameter and NO arithmetic calculation}. We desire to know whether ViT can retain the good performance under this extreme case.

Conceptually, this zero-parameter alternative must rely on the handcrafted rule to model spatial relationships. In this work, we revisit the shift operation, which we believe is one of the simplest spatial modeling module. As depicted in Figure \ref{fig:architecture}, the standard ViT building block consists of two parts: the attention layer and the feed-forward network (FFN). We replace the former attention layer with a shift operation, while keeping the latter FFN part untouched. Given an input feature, the proposed building block will first shift a small portion of the channels along four spatial directions, namely left, right, top, and down. As such, the information of neighboring features is explicitly mingled by the shifted channels. Then, the subsequent FFN performs channel-wise mixing to further fuse the information from neighbors.

Based on this shift building block, we construct a ViT-like backbone network, namely ShiftViT. Surprisingly, this backbone can also work well for the mainstream visual recognition tasks. The performance is on par with or even better than the strong Swin Transformer baseline. Concretely, within the same computational budgets as Swin-T model, our ShiftViT achieves a top-1 classification accuracy of 81.7\% (against Swin-T's 81.3\%) on ImageNet dataset. For the dense prediction task, it attains a mean average precision (mAP) score of 45.7\% (against Swin-T's 43.7\%) on COCO detection dataset, and a mean IoU (mIoU) score of 46.3\% (against Swin-T's 44.5\%) on ADE20k segmentation dataset.

Since the shift operation is already the simplest spatial modelling module, the excellent performance must come from the remaining components, e.g., the linear layers and the activation function in FFN. These components are less studied in existing work, because they look trivial. However, to further demystify the reasons why ViT works, we argue that we should pay more attentions to these components, instead of just focusing on the attention mechanism. We hope our work can shed a new light on the ViT research. As a summary, the contributions of this work are two folds:

\begin{itemize}
\item We present a ViT-like backbone, where the vanilla attention layer is replaced by an extremely simple shift operation. The proposed model can achieve an even better performance than Swin Transformer.
\item We analyze the reasons behind the success of ViTs. It hints that the attention mechanism might not be the vital factor that makes ViT work. We should take the remaining components seriously in the future study of ViTs.
\end{itemize}

\section{Related Work}

\subsection{Attention and Vision Transformers}

Transformer architecture \cite{AttentionAllYouNeed} is first introduced in the area of natural language processing (NLP). It solely adopts attention mechanism to build the connections between different language tokens. Thanks to the great performance, Transformers have rapidly dominated the NLP area and become the \textit{de facto} standard.

Inspired by the successful application in NLP, attention mechanism has also received increasing interests from the computer vision community. The early explorations can be roughly divided into two categories. On the one hand, some literature considers attention as a plug-and-play module, which can be seamlessly integrated into the existing CNN architectures. The representative work includes non-local network \cite{NonLocal}, relation network \cite{RelationNet}, and CCNet \cite{CCNet}. On the other hand, some pieces of work aim to substitute all convolution operations with the attention mechanism, such as local relation network \cite{LocalRelation} and self-attention network \cite{SANet}. 

Although these two kinds of work have shown promising results, they are still built on the CNN architecture. ViT \cite{ViT} is the pioneering work that leverages a pure transformer architecture for visual recognition tasks. Thanks to its impressive performance, the community recently bursts out a rising wave of research on vision transformers. Along this line of research, the main focus is to improve the attention mechanism, so that it can satisfy the intrinsic properties of visual signals. For example, MSViT \cite{MSViT} builds hierarchical attention layers to obtain multi-scale features. Swin Transformers \cite{Swin} introduces a locality constrain into its attention mechanism. The related efforts also include pyramid attention \cite{PVT}, local-global attention \cite{LocalGlobalTransformer}, cross attention \cite{CrossViT}, to name a few.

Unlike the particular interests in attention mechanism, the remaining components of ViT are less studies. DeiT \cite{DeiT} has setup a standard training pipeline for vision transformers. Most follow-up work inherits its setting, and only make some modifications on the attention mechanism. Our work also follows this paradigm. However, the goal of this work is not to complex the design of attention. On the contrary, we aim to show that the attention mechanism might not be the critical part of making ViTs work. It can be even replaced by an extremely simple shift operation. We hope these results can inspire researchers to rethink the role of attention mechanism. 

\begin{figure*}[]
\centering
\includegraphics[width=\linewidth]{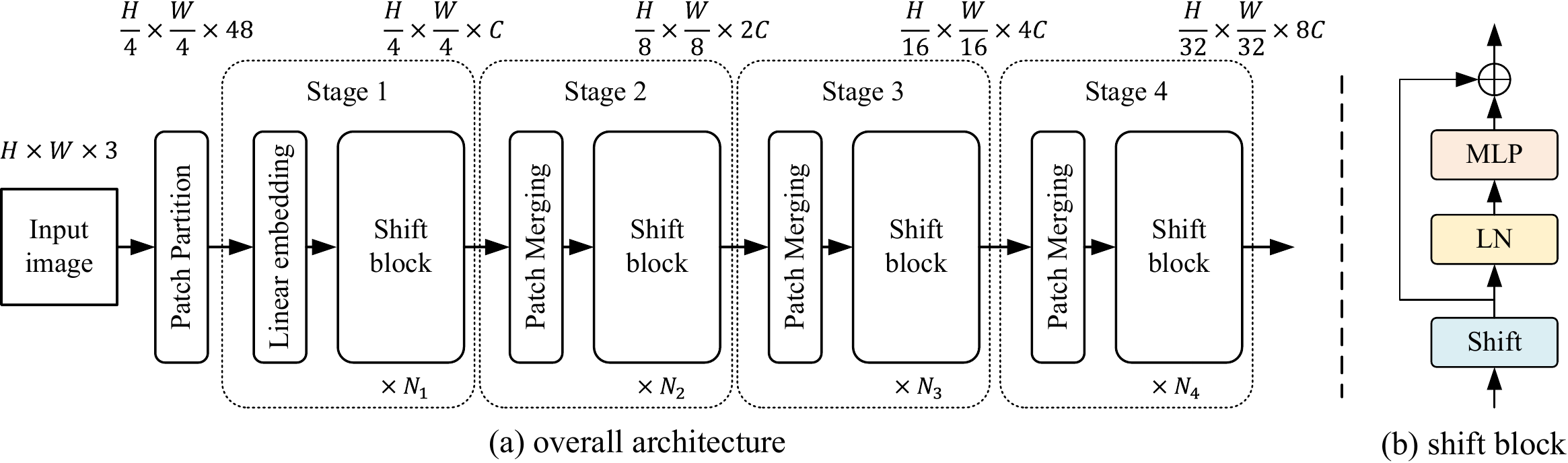}
\caption{(a) The overall architecture of our ShiftViT. We follow Swin Transformer \cite{Swin} to build hierarchical representations. (b) The detail design of a shift block. We only use a simple shift operation to model spatial relationships.}
\label{fig:overview}
\end{figure*}

\subsection{MLP Variants}

Our work is related to the recent multi-layer-perceptron (MLP) variants. Specifically, MLP variants propose to extract image features through a pure MLP-like architecture. They also jump out of the attention-based framework in ViT. For example, instead of using the self-attention matrix, MLP-Mixer \cite{MLPMixer} introduces a token-mixing MLP to directly connect all spatial locations. It eliminates the dynamic property of ViT, but without losing accuracy. The follow-up work investigates more MLP designs, like the spatial gating unit \cite{gMLP} or cyclic connection \cite{CycleMLP}.

Our ShiftViT can be also categorized into the pure MLP architecture, where the shift operation is viewed as a special token-mixing layer. Compared with the existing MLP work, our shift operation is even much simpler, since it contains no parameter and no FLOP. Moreover, the vanilla MLP variants fail to handle variable input size because of the fixed linear weights. Our shift operation overcomes this obstacle and therefore make the backbone feasible for more vision tasks like object detection and semantic segmentation.


\subsection{Shift Operation}

Shift operation is not new in computer vision. As early as in 2017, it was proposed to be an efficient alternative to the spatial convolution operation \cite{ShiftResNet}. Concretely, it uses a sandwich-like architecture, two $1\times1$ convolutions and a shift operation, to approximate a $K \times K$ convolution. In the follow-up work, the shift operation is further extended into different variants, such as active shift \cite{ActiveShift}, sparse shift \cite{SparseShift} and partial shift \cite{TSM}.

In this work, we adopt the partial shift operation \cite{TSM}. It is notable that the goal of this work is not to present a novel operation. Instead of that, we integrate the existing shift operation with the popular ViT to verify the effectiveness of attention mechanism. The similar vision are shared with the concurrent work ShiftMLP \cite{ShiftMLP} and AS-MLP \cite{ASMLP}, but the design details are quite different. Their building blocks are more complex, which involve some auxiliary layers like pre-transformation and post-transformation.

\section{Shift Operation Meets Vision Transformer}

\subsection{Architecture Overview}

For a fair comparison, we follow the architecture of Swin Transformer \cite{Swin}. The architecture overview is illustrated in Figure \ref{fig:overview} (a). Specifically, given an input image of shape $H\times W \times 3$, it first splits the images into non-overlapping patches. The patch size is $4\times 4$ pixels. Therefore, the output of patch partition is is $\frac{H}{4}\times\frac{W}{4}$ tokens, where each token has a channel size of $48$. 

The modules followed by can be divided into 4 stages. Each stage contains two parts: embedding generation and stacked shift blocks. For the embedding generation of the first stage, a linear projection layer is used to map each token into an embedding of channel size $C$. For the rest stages, we merge neighbouring patches through the convolution with a kernel size of $2\times2$. After patch merging, the spatial size of the output is half down-sampled, while channel size is twice the input, i.e., from $C$ to $2C$.

The stacked shift block is built by some repeated basic units. The detail design of each shift block is shown in Figure \ref{fig:overview} (b). It composes of a shift operation, a layer normalization and a MLP network. This design is almost the same as the standard transformer block. The only difference is that we use a shift operation rather than a attention layer. For each stage, the number of shift blocks can be various, which is denoted as $N_1, N_2, N_3, N_4$ respectively. In out implementation, we carefully choose the value of $N_{i}$ so that the overall model share a similar number of parameters with the baseline Swin Transformer model.

\subsection{Shift Block}

The detail architecture of our shift block is depicted in Figure \ref{fig:overview} (b). Specifically, this block consists of three sequentially-stacked components: shift operation, layer normalization and MLP network.

Shift operation has been well studied in CNNs. It can have many design choices, such as active shift \cite{ActiveShift} and sparse shift \cite{SparseShift}. In this work, we follow the partial shift operation in TSM \cite{TSM}. The illustration is presented in Figure \ref{fig:architecture} (b). Given an input tensor, a small portion of channels will be shifted along 4 spatial directions, namely left, right, top, and down, while the remaining channels keep unchanged. After shifting, the out-of-scope pixels are simply dropped and the vacant pixels are zero padded. In this work, the shift step is set to 1 pixel.

Formally, we assume that the input feature $\mathbf{z}$ is of shape $H\times W\times C$, where $C$ is the number of channels, $H$ and $W$ are spatial height and width, respectively. The output feature $\hat{\mathbf{z}}$ has the same shape as input. It can be written as:

\begin{equation*}
\small
\begin{split}
\hat{\mathbf{z}}[0:H, 1:W, 0:\gamma C]&\leftarrow \mathbf{z}[0:H, 0:W-1, 0:\gamma C] \\
\hat{\mathbf{z}}[0:H, 0:W-1, \gamma C:2\gamma C]&\leftarrow \mathbf{z}[0:H, 1:W, \gamma C:2\gamma C] \\
\hat{\mathbf{z}}[0:H-1, 0:W, 2\gamma C:3\gamma C] & \leftarrow \mathbf{z}[1:H, 0:W, 2\gamma C:3\gamma C] \\
\hat{\mathbf{z}}[1:H, 0:W, 3\gamma C:4\gamma C]&\leftarrow \mathbf{z}[0:H-1, 0:W, 3\gamma C:4\gamma C] \\
\hat{\mathbf{z}}[0:H, 0:W, 4\gamma C:C]&\leftarrow \mathbf{z}[0:H, 0:W, 4\gamma C:C] \\
\end{split}
\end{equation*}

\noindent where $\gamma$ is a ratio factor to control how many percentages of channels will be shifted. In most experiments, the value of $\gamma$ is set to $\frac{1}{12}$.

It is notable that shift operation does not hold any parameter or arithmetic calculation. The only implementation is memory copying. Therefore, shift operation is highly efficient and it is very easy to implement. The pseudo code is presented in Algorithms \ref{alg:code}. Compared with the self-attention mechanism, shift operation is clean, neat, and more friendly to deep learning inference library like TensorRT.

The rest of the shift block is the same as the standard building block of ViT. The MLP network has two linear layers. The first one increases the channel of the input feature to a higher dimension, e.g., from $C$ to $\tau C$. Then the second linear layer projects the high-dimensional feature into the original channel size of $C$. Between these two layers, we adopt GELU as the non-linear activation function.

\subsection{Architecture Variants}

For a fair comparison with the baseline Swin Transformer, we also build multiple models with various number of parameters and computational complexity. Specifically, we introduce Shift-T(iny), Shift-S(mall), Shift-B(ase) variants \footnote{For simplification, we ignore the suffix of ``ViT'' and use Shift-T to denote ShiftViT-T in this work.}, which is corresponded to Swin-T, Swin-S and Swin-B, respectively. Shift-T is the smallest one, which shares a similar size with Swin-T and ResNet-50. Another two variants, Shift-S and Shift-B, are roughly $2\times$ and $4\times$ more complex than ShiftViT-T. The detail configurations of basic embedding channels $C$ and number of blocks $\{N_{i}\}$ are presented as following:

\begin{itemize}
\item Shift-T: $C=96$, $\{N_{i}\}=\{6, 8, 18, 6\}$, $\gamma=1/12$
\item Shift-S: $C=96$, $\{N_{i}\}=\{10, 18, 36, 10\}$, $\gamma=1/12$
\item Shift-B: $C=128$, $\{N_{i}\}=\{10, 18, 36, 10\}$, $\gamma=1/16$
\end{itemize}

Beside the model size, we also have a closer look at the model depth. In our proposed model, nearly all parameters are concentrated in the MLP part. Therefore, we can control the expand ratio of MLP $\tau$ to obtain a deeper network depth. If not specified, the expand ratio $\tau$ is set to 2. We have an ablation analysis to show that the deeper model achieve a better performance.


\begin{algorithm}[t!]
\caption{Pytorch-like pseudo code of shift}
\label{alg:code}
\centering
\definecolor{codeblue}{rgb}{0.25,0.5,0.5}
\lstset{
  backgroundcolor=\color{white},
  basicstyle=\fontsize{7.2pt}{7.2pt}\ttfamily\selectfont,
  columns=fullflexible,
  breaklines=true,
  captionpos=b,
  commentstyle=\fontsize{7.2pt}{7.2pt}\color{codeblue},
  keywordstyle=\fontsize{7.2pt}{7.2pt},
  escapeinside={(*@}{@*)}
}
\begin{lstlisting}[language=python]
def shift(feat, gamma=1/12):
    # feat is a tensor with a shape of 
    # [Batch, Channel, Height, Width]
    B, C, H, W = feat.shape
    g = int(gamma * C)
    out = zeros_like(feat)
    # spatially shift
    out[:, 0*g:1*g, :, :-1] = x[:, 0*g:1*g, :, 1:]  
    out[:, 1*g:2*g, :, 1:] = x[:, 1*g:2*g, :, :-1]
    out[:, 2*g:3*g, :-1, :] = x[:, 2*g:3*g, 1:, :]
    out[:, 3*g:4*g, 1:, :] = x[:, 3*g:4*g, :-1, :]
    # remaining channels
    out[:, 4*g:, :, :] = x[:, 4*g:, :, :]
    return out
    
\end{lstlisting}
\end{algorithm}

\section{Experiments}

\subsection{Implementation Details}

We conduct experiments on three mainstream visual recognition benchmarks: image classification on ImageNet-1k dataset \cite{ImageNet}, object detection on COCO dataset \cite{COCO} and semantic segmentation on ADE20k dataset \cite{ADE20K}. 

For image classification task, we exactly follow the protocol as in Swin Transformer \cite{Swin}. An average pooling layer and a linear classification layer are appended after the backbone network. All the parameters are randomly initialized and trained for 300 epochs with an AdamW optimizer. The learning rate starts from 0.001 and gradually decay to 0 with a cosine schedule. We include all data augmentations and regularization tricks as in Swin Transformer \cite{Swin}. The batch size is set to 1024.

For object detection task, there exists many off-the-shelf detection frameworks, such as Faster R-CNN, Mask R-CNN and RetinaNet. For a fair comparison with other methods, we follow the common practice of using Mask R-CNN and Cascade Mask R-CNN. In such detection frameworks, the backbone is our proposed Shift network, while the rest of components like FPN and detection head remain the same. We initialize the backbone with pretrained weights of the ImageNet-1k classifier. The training duration lasts for 12 epochs (denoted as $1\times$ schedule) or 36 epochs (denoted as $3\times$ schedule). The optimizer is AdamW, with an initial learning rate 0.0001. The batch size is 16. During training period, we utilize the multi-scale training trick, i.e., the shorter side of the input image is resized into a range from 480 pixels to 800 pixels. We report the mean average precision (mAP) metrics on the validation set of COCO dataset.

For semantic segmentation task, we evaluate our method on ADE20K dataset, which contains 20K images for training and 2K images for validation. In these experiments, the base segmentation framework is UperNet. The model is trained on the training set of ADE20K and the evaluation metric is the mean IoU (mIoU) score on the validation set. Similar to the setting of object detection, our Shift backbones are also pretrained on ImageNet-1k. The rest of settings are same as Swin-Transformer. The training batch size is 16 and we train the model for 160k iterations. For the comparison with the state-of-the-arts, we adopt the multi-scale testing strategy.

\begin{table*}[]
\centering
\caption{Comparison with the baseline Swin Transformer on three mainstream tasks: image classification, object detection and semantic segmentation. The suffix \texttt{/light} denotes the lightweight version of our ShiftViT, where we only replace attention layers with the shift operation and keep remaining parts unchanged. The throughput speed is evaluated on a single NVidia GTX1080-Ti GPU.The \textcolor{Highlight}{gree} and \textcolor{Bad}{gray} colors indicate the gain and loss, respectively.}

\label{table:baseline}
\setlength\tabcolsep{5pt}
\begin{tabular}{@{}l|c|ccl|ll|ll|l@{}}
\toprule
\multirow{3}{*}{Model} & \multirow{3}{*}{\begin{tabular}[c]{@{}c@{}}Param\\ (M)\end{tabular}} & \multicolumn{3}{c|}{ImageNet} & \multicolumn{4}{c|}{COCO} & \multicolumn{1}{c}{ADE20k} \\
 &  & FLOPs & Speed & \multicolumn{1}{c|}{Top-1} & \multicolumn{2}{c}{Mask R-CNN $1\times$} & \multicolumn{2}{c|}{Mask R-CNN $3\times$}  & \multicolumn{1}{c}{UpperNet} \\
 &  & (G) & (FPS) & \multicolumn{1}{c|}{Acc.(\%)} & \multicolumn{1}{c}{AP$^b$} & \multicolumn{1}{c}{AP$^m$} & \multicolumn{1}{c}{AP$^b$} & \multicolumn{1}{c|}{AP$^m$} & \multicolumn{1}{c}{mIoU} \\ \midrule
ResNet-50 & 26 & 4.1 & 676 & 76.1 & 38.0 & 34.4 & 41.0 & 37.1 & - \\ \midrule
Swin-T & 29 & 4.5 & 356 & 81.3 & 43.7 & 39.5 & 46.0 & 41.6 & 44.5 \\
Shift-T/light & 20 & 3.0 & 790 & 79.4 & 41.3 & 38.0 & 43.2 & 39.2 & 42.6 \\
Shift-T & 29 & 4.5 & 396 & 81.7 \textcolor{Highlight}{(+0.4)} & 45.4 \textcolor{Highlight}{(+1.7)} & 40.9 \textcolor{Highlight}{(+1.4)} & 47.1 \textcolor{Highlight}{(+1.1)} & 42.3 \textcolor{Highlight}{(+0.7)} & 46.3 \textcolor{Highlight}{(+1.8)}  \\ \midrule
Swin-S & 50 & 8.7 & 217 & 83.0 & 46.4 & 41.7 & 48.5 & 43.3 & 47.6 \\
Shift-S/light & 34 & 5.7 & 457 & 81.6 & 44.8 & 40.4 & 46.0 & 41.1 & 45.4 \\
Shift-S & 50 & 8.8 & 215 & 82.8 \textcolor{Bad}{(-0.2)} & 47.2 \textcolor{Highlight}{(+0.8)} & 42.2 \textcolor{Highlight}{(+0.5)} & 48.6 \textcolor{Highlight}{(+0.1)} & 43.4 \textcolor{Highlight}{(+0.1)} & 47.8 \textcolor{Highlight}{(+0.2)}  \\ \midrule
Swin-B & 88 & 15.4 & 158 & 83.5 & 46.9 & 42.1 & 48.7 & 43.4 & 48.1 \\
Shift-B/light & 60 & 10.2 & 312 & 82.3 & 45.7 & 41.0 & 46.0 & 41.2 & 45.8 \\
Shift-B & 89 & 15.6 & 154 & 83.3 \textcolor{Bad}{(-0.2)} & 47.7 \textcolor{Highlight}{(+0.8)} & 42.7 \textcolor{Highlight}{(+0.6)} & 48.0 \textcolor{Bad}{(-0.7)} & 42.8 \textcolor{Bad}{(-0.6)} & 47.9 \textcolor{Bad}{(-0.2)}  \\ \bottomrule
\end{tabular}
\end{table*}

\begin{table}[]
\centering
\caption{Comparison with state-of-the-art methods on the ImageNet-1k classification task.}
\label{table:classification}
\setlength\tabcolsep{4pt}
\begin{tabular}{@{}l|c|c|c|c@{}}
\toprule
\multirow{2}{*}{Model} & Input & \# Params & FLOPs & Top-1 \\
 & resolution & (M) & (B) & Acc. (\%) \\ \midrule
 \multicolumn{5}{c}{CNN-based} \\ \midrule
 RegNetY-4G & $224^2$ & 21 & 4.0 & 80.0 \\
 RegNetY-8G & $224^2$ & 39 & 8.0 & 81.7 \\
 RegNetY-16G & $224^2$ & 84 & 16.0 & 82.9 \\
 EfficientNet-B4 & $380^2$ & 19 & 4.2 & 82.9 \\
 EfficientNet-B5 & $456^2$ & 30 & 9.9 & 83.6 \\
 EfficientNet-B6 & $528^2$ & 43 & 19.0 & 84.0 \\ \midrule
 \multicolumn{5}{c}{ViT-based and MLP-based} \\ \midrule
 DeiT-S & $224^2$ & 22 & 4.6 & 79.8 \\
 DeiT-B & $224^2$ & 86 & 17.5 & 81.8 \\
 PVT-S & $224^2$ & 25 & 3.8 & 79.8 \\
 PVT-L & $224^2$ & 61 & 9.8 & 81.7 \\
 Swin-T & $224^2$ & 29 & 4.5 & 81.3 \\
 Swin-S & $224^2$ & 50 & 8.7 & 83.0 \\
 Swin-B & $224^2$ & 88 & 15.4 & 83.5 \\ \midrule
 MLP-Mixer-B/16 & $224^2$ & 79 & - & 76.4 \\
 gMLP-S & $224^2$ & 20 & 4.5 & 79.4 \\
 gMLP-B & $224^2$ & 73 & 15.8 & 81.6 \\ 
 S$^2$-MLP-D & $224^2$ & 71 & 14.0 & 80.0 \\
 S$^2$-MLP-W & $224^2$ & 51 & 10.5 & 80.7 \\
 AS-MLP-T & $224^2$ & 28 & 4.4 & 81.3 \\
 AS-MLP-S & $224^2$ & 50 & 8.5 & 83.1 \\
 AS-MLP-B & $224^2$ & 88 & 15.2 & 83.3 \\ \midrule
  \multicolumn{5}{c}{Ours} \\ \midrule
 Shift-T & $224^2$ & 28 & 4.4 & 81.7 \\
 Sfhit-S & $224^2$ & 50 & 8.5 & 82.8 \\
 Sfhit-B & $224^2$ & 88 & 15.2 & 83.3 \\ \bottomrule 
\end{tabular}
\end{table}

\subsection{Comparison with Baseline}

The goal of this work is to demystify the role of attention mechanism and explore whether it can be replaced by an extremely simple shift operation. Concretely, our proposed backbones are based on the architecture of Swin Transformer, which is one of the most representative ViT variants. We therefore consider Swin Transformer as the baseline model, and compare our ShiftViT to it.

For an apple-to-apple comparison, we first build a lightweight version of ShiftViT. It is nearly the same as the Swin Transformer counterpart, except that the attention layers are substituted by the shift operations. We denote this backbone with a suffix \texttt{/light}, because replacing attention with shift will lead to a reduction in parameters and FLOPs. The experimental results are presented in Table \ref{table:baseline}. We exhaustively compare all variants in three different sizes. The results show that the shift operation is weaker than the attention mechanism, because it does not contain any learnable parameter or arithmetic calculation. For example, the Shift-T/light model has only 20M parameters and 3.0 FLOPs, which are nearly 33\% less than the Swin-T model. Therefore, there is no wonder that its performance is marginally worse than the baseline. Despite the relative gap to the baseline, it is worth noting that the absolute accuracy of the lightweight ShiftViT is not bad. Compared with the typical ResNet-50 backbone, Shift-T/light is more powerful and more efficient.

To remedy the complexity gap between shift operation and attention mechanism, we can adopt more building blocks in ShiftViT to make sure it has a similar number of parameters with the Swin baseline. In such fair comparisons, our models achieve even better results than Swin-Transformer. For the small-size models, our Shift-T backbone attains an mAP score of 45.4\% on COCO and an mIoU score of 46.3\% on ADE20k, which outperform the Swin-T backbone by a remarkable margin. For the large-size models, ShiftViT seems to be saturated. But the performance is still on par with the Swin baseline.

Although the shift operation is weaker than the attention mechanism in spatial modelling, its simple architecture allows the network to grow deeper. As such, the weakness of the shift operation is greatly alleviated. Within the same computational budget, the overall performance of ShiftViT is comparable to the attention-based Swin Transformer. These experiments prove that the attention mechanism might not be necessary for ViTs. Even an extremely simple operation can achieve the similar results.

\subsection{Comparison with State-of-the-Art}

To further demonstrate the effectiveness, we compare ShiftViT backbones with existing state-of-the-art methods. For image classification task on ImageNet-1k, our proposed models are compared to three different types of models, namely CNN, ViT and MLP. The results are detailed in Table \ref{table:classification}. Overall, our method can achieve a comparable performance with the state-of-the-arts. For ViT-based and MLP-based methods, the best performances are around 83.5\% top-1 accuracy, while our model achieves an accuracy of 83.3\%. For CNN-based methods, our model is slightly worse than EfficientNet series, but the comparison is not fully fair because EfficientNet takes a larger input size. 

Another interesting thing is the comparison with two concurrent work S$^2$-MLP \cite{ShiftMLP} and AS-MLP \cite{ASMLP}. These two pieces of work share the similar idea on shift operation , but they introduce some auxiliary modules into the building block, e.g., the pre- and post-projection layers. In Table \ref{table:classification}, our performances are slightly better than these two work. It justifies our design choice that building backbone solely with a simple shift operation is good enough.

Beside the classification task, the similar performance tread can be also observed in the object detection task and semantic segmentation task. It is notable that some ViT-based and MLP-based methods cannot be easily extended to such dense prediction tasks, because the high-resolution inputs yield unaffordable computational burdens. Our method does not suffer from this obstacle thanks to the high efficiency of shift operation. As shown in Table \ref{table:detection} and Table \ref{table:segmentation}, the advantages of our ShiftViT backbones are clear. Shift-T attains an mAP score of 47.1 on object detection and an mIoU score of 47.8 on semantic segmentation, which outperform other methods by a considerable margin. 

\begin{table}[]
\centering
\caption{Comparison with state-of-the-art methods on the COCO object detection task. Following the common practice, we couple the backbones with two detection frameworks, namely Mask R-CNN and Cascade Mask R-CNN.}
\label{table:detection}
\begin{tabular}{l|cc|cc}
\toprule
Backbone       & Params (M) & FLOPs (G) & AP$^{b}$ & AP${^m}$ \\ \midrule
\multicolumn{5}{c}{Mask R-CNN $3\times$} \\ \midrule
Res-50         & 44     & 260   & 41.0     & 37.1     \\ 
PVT-S          & 44     & 245   & 43.0     & 39.9     \\
AS-MLP-T       & 48     & 260   & 46.0     & 41.5     \\
Swin-T         & 48     & 264   & 46.0     & 41.6     \\
\rowcolor{Bg} Shift-T   & 48    & 265       & 47.1     & 42.3     \\ \midrule
Res-101        & 63     & 336   & 42.8     & 38.5     \\ 
PVT-M          & 64     & 302   & 44.2     & 40.5     \\
AS-MLP-S       & 69     & 346   & 47.8     & 42.9     \\
Swin-S         & 69     & 354   & 48.5     & 43.3     \\
\rowcolor{Bg} Shift-S        & 70     & 350  & 48.6     & 43.4     \\ \midrule
\multicolumn{5}{c}{Cascade Mask R-CNN $3\times$} \\ \midrule
Res-50         & 82     & 739   & 46.3     & 40.1     \\ 
AS-MLP-T       & 86     & 745   & 50.1     & 43.5     \\
Swin-T         & 86     & 739   & 50.4     & 43.7     \\
\rowcolor{Bg} Shift-T        & 86     & 743      & 50.3     & 43.4     \\ \midrule
ResX-101       & 101    & 819   & 48.1     & 41.6     \\
AS-MLP-S       & 107    & 824   & 51.1     & 44.2     \\
Swin-S         & 107    & 838   & 51.8     & 44.7     \\
\rowcolor{Bg} Shift-S        & 107    & 827      & 50.9     & 44.0     \\ \bottomrule
\end{tabular}
\end{table}

\begin{table}[]
\centering
\caption{Comparison with state-of-the-art methods on the ADE20k semantic segmentation task. We report the mIoU metrics on the validation set.}
\label{table:segmentation}
\begin{tabular}{l|l|cc|c}
\toprule
\multirow{2}{*}{Method} & \multirow{2}{*}{Backbone}  & Params & FLOPs & val \\ 
& & (M) & (G) & mIoU \\ \midrule

DANet & ResNet-101 & 69 & 1119 & 45.2 \\ 
DNL & ResNet-101 & 69 & 1249 & 46.0 \\
DeepLabV3 & ResNet-101 & 63 & 1021 & 44.1 \\
OCRNet & ResNet-101 & 89 & 1381 & 44.9 \\
DeepLabV3 & ResNeSt-101 & 66 & 1051 & 46.9 \\
DeepLabV3 & ResNeSt-200 & 88 & 1381 & 48.4 \\
OCRNet & HRNet-w64 & 71 & 664 & 45.7 \\ \midrule
UperNet & ResNet-101 & 89 & 1029 & 44.9 \\
UperNet & Swin-T & 60 & 945 & 45.8 \\
UperNet & AS-MLP-T & 60 & 937 & 46.5 \\
\rowcolor{Bg} UperNet & Shift-T & 60 & 942 & 47.8\\ \midrule
UperNet & Swin-S & 81 & 1038 & 49.5 \\
UperNet & AS-MLP-S & 81 & 1024 & 49.2 \\
\rowcolor{Bg} UperNet & Shift-S & 81 & 1029 & 49.6\\ \midrule
UperNet & Swin-B & 121 & 1188 & 49.7 \\
UperNet & AS-MLP-B & 121 & 1166 & 49.5 \\
\rowcolor{Bg} UperNet & Shift-B & 121 & 1174 & 49.2 \\
\bottomrule

\end{tabular}
\end{table}

\subsection{Ablation Analysis}

In this section, we aim to explore what factors contribute to the good performance of ShiftViT. We first analyze the impact of two hyper-parameters in ShiftViT. Then, we dive into the training scheme of ViT series.

\subsubsection{Expand ratio of MLP}

The previous experiments have justified our design principle, i.e., a great model depth can remedy the weakness of each building block. Generally, there exists a trade-off between the model depth and the complexity of building blocks. With a fixed computational budget, a lightweight building block can enjoy a deeper network architecture.

To further investigate this trade-off, we present some ShiftViT models with different depths. For ShiftViT, most parameters exist in the MLP part. We can change the expand ratio of MLP $\tau$ to control the model depth. As shown in Table \ref{table:expand_ratio}, we choose Shift-T as our baseline model. We explore the expand ratio $\tau$ within a range from 1 to 4. It is worth noting that the parameters and FLOPs for different entries are almost the same. 
From Table \ref{table:expand_ratio}, we can observe a trend that a deeper model results in a better performance. When the depth of ShiftViT increases to 225, it outperforms the 57-layer counterpart by 0.5\%, 1.2\% and 2.9\% absolute gains on classification, detection and segmentation, respectively. This trend supports our conjecture that a powerful-and-heavy module, like attention, may not be the optimal choice for backbone. We hope it can help the future work to rethink such trade-off when designing backbones.

\begin{table}[]
\centering
\caption{Ablation analysis on the expand ratio of MLP. The first row shows the Swin-T baseline. The row with \textcolor{Blue}{blue} background denotes the default setting in our experiments. All entries share the same number of parameters and FLOPs.}
\label{table:expand_ratio}
\begin{tabular}{c|c|c|cc|c}
\toprule
Expand & \multirow{2}{*}{Depth}  & ImgNet & \multicolumn{2}{c|}{COCO} & ADE20k \\ 
Ratio & & Acc. (\%) & AP$^{b}$ & AP$^{m}$ & mIoU \\ \midrule
Swin & 48 & 81.3 & 43.7 & 39.5 & 44.5 \\ \midrule
4 & 57 & 81.3 & 44.0 & 39.8 & 44.4 \\
3 & 75 & 81.5 & 44.4 & 40.2 & 45.5 \\
\rowcolor{Bg} 2 & 114 & 81.7 & 45.4 & 40.9 & 46.3 \\
1 & 225 & 81.8 & 45.2 & 40.6 & 47.3\\
\bottomrule
\end{tabular}
\end{table}

\subsubsection{Percentage of shifted channels} 
\begin{figure}[t]
\centering
\includegraphics[width=0.9\linewidth]{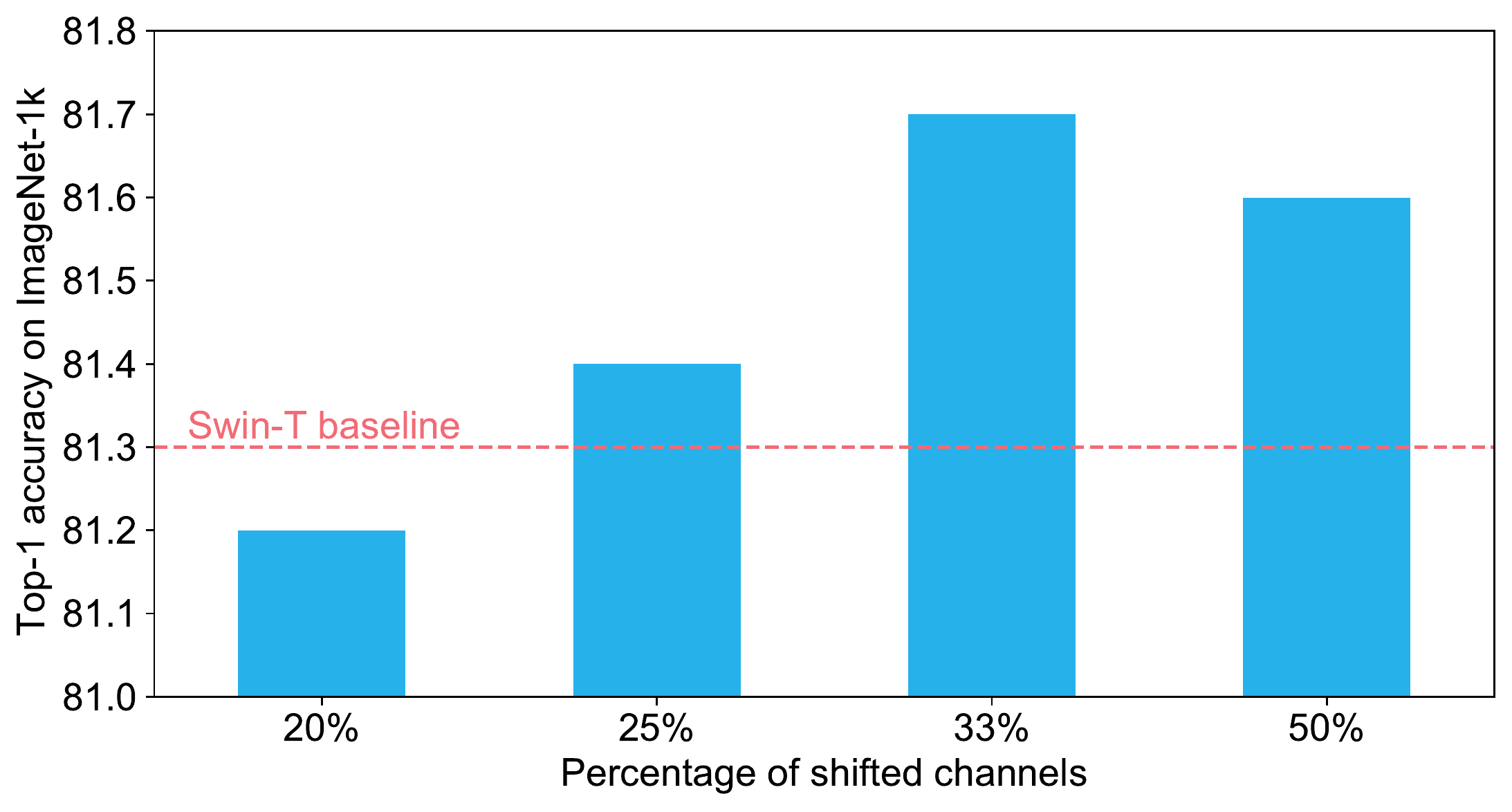}
\caption{Ablation analysis on the percentage of shifted channels. We plot the top-1 classification accuracy on ImageNet-1k. The \textcolor{red}{red} line indicates Swin-T baseline.}
\label{fig:acc_vs_percentage}
\end{figure}

The shift operation has only one hyper-parameter, namely the percentage of shifted channels. By default, it is set to 33\%. In this section, we explore some other settings. Specifically, we set the percentage of shifted channels to 20\%, 25\%, 33\% and 50\%, respectively. The results are presented in Figure \ref{fig:acc_vs_percentage}. It shows that the final performance is not very sensitive to this hyper-parameter. Shifting 25\% of channels only results in 0.3\% absolute loss compared to the best setting. Within the reasonable range (from 25\% to 50\%), all the settings achieve a better accuracy than the Swin-T baseline. 

\subsubsection{Shifted pixels} In the shift operation, a small portion of channels are shifted by one pixel along four directions. To have a comprehensive exploration, we also try different shifted pixels. When the shifted pixel is zero, i.e., no shifting happens, the top-1 accuracy on the ImageNet dataset is only 72.9\%, which is significantly lower than our baseline (81.7\%). This is not surprising because no shifting means there is no interaction between different spatial location. Besides, if we shift two pixels in the shift operation, the model achieves 80.2\% top-1 accuracy on ImageNet, which is also slightly worse than the default setting. 

\begin{table}[]
\centering
\caption{Ablation analysis on the typical configurations of CNNs and ViTs. We gradually transfer the training configuration from the CNN's setting to the ViT's setting, and investigate how these factors influence the model performances.}
\label{table:vit_configuration}
\begin{tabular}{cccc|c}
\toprule
 SGD & ReLU & BN & 90ep & ImageNet \\
 $\downarrow$& $\downarrow$ & $\downarrow$ & $\downarrow$ & Top-1 Acc.\\
AdamW & GELU & LN & 300ep & (\%)\\ \midrule
 & & & & 76.4 \\
 \checkmark & & & & 77.9 \\
 \checkmark & \checkmark & & & 78.5 \\
 \checkmark & \checkmark & \checkmark & & 78.4 \\
 \checkmark & \checkmark & \checkmark & \checkmark & 81.7 \\
\bottomrule
\end{tabular}
\end{table}

\subsubsection{ViT-style training scheme} Shift operation has been well studied in CNNs. However, the previous work does not show the impressive performance as ours. Shift-ResNet-50 \cite{ShiftResNet} only achieve an accuracy of 75.6\% on ImageNet, which is far behind our 81.7\% accuracy. This gap raise a natural concern about what makes good for our ShiftViT.

We suspect the reason might lie in the ViT-style training scheme. Specifically, most existing ViT variants follow the setting as in DeiT \cite{DeiT}, which is quite different from the standard pipeline of training CNNs. For example, ViT-style scheme adopts AdamW optimizer and the training duration lasts for 300 epochs on ImageNet. As a comparison, CNN-style scheme prefers SGD optimizer and the training schedule is usually 90 epochs only. Since our model inherit the ViT-style training scheme, it is interesting to see how such differences affect the performance.

Due to the resource limitation, we cannot fully align all settings between ViT-style and CNN-style. Therefore, we pick four important factors that we believe can bring some insights, i.e. optimizer, activation function, normalization layer and training schedule. From Table \ref{table:vit_configuration}, we can observe that such factors can significantly influence the accuracy, especially the training schedule. These results shows that the good performance of ShiftViT is partly brought by the ViT-style training scheme. Similarly, the success of ViT may be also related to its special training scheme. We should take it seriously in the future study of ViTs.



\section{Conclusion}

In this work, we move a small step toward demystifying the essential reason why ViT works. The experiments show that the attention mechanism might not be the vital factor for the success of ViT. We can even use an extremely simple shift operation to replace the attention layer. The proposed backbone, namely ShiftViT, can work as well as the Swin Transformer baseline. Since the shift operation is already the simplest spatial modelling module, we argue that the good performance must come from the remaining components of ViT, e.g., the FFN and the training scheme. In future work, we plan to have more analysis on such factors and investigate more ViT variants.

\bibliography{aaai22}
\end{document}